\RequirePackage{fix-cm}
\documentclass[twocolumn]{svjour3}          
\smartqed
\usepackage[utf8]{inputenc}
\usepackage{graphicx}
\usepackage{mathptmx}
\usepackage{amsmath}
\usepackage{amsfonts}
\usepackage{hyphenat}
\usepackage{array}
\usepackage{xspace}
\usepackage{csquotes}
\usepackage{tabularx}
\usepackage[]{algorithmicx}
\usepackage[]{algpseudocode}
\usepackage{float}

\makeatletter
\def\cl@chapter{\@elt {theorem}}
\makeatother
\usepackage{hyperref}
\usepackage{cleveref}

\usepackage{tikz}
\usepackage{pgfplots}
\usepackage{numprint}

\newfloat{algorithm}{h}{lop}
\floatname{algorithm}{\small\textbf{Algorithm}}

\makeatletter
\algrenewcommand\ALG@beginalgorithmic{\ttfamily\small}
\makeatother

\makeatletter
\DeclareRobustCommand\onedot{\futurelet\@let@token\@onedot}
\def\@onedot{\ifx\@let@token.\else.\null\fi\xspace}

\def\eg{\emph{e.g}\onedot} 
\def\ie{\emph{i.e}\onedot}

\def\wrt{w.r.t\onedot} 
\def\etal{\emph{et al}\onedot}
\makeatother

\newcommand\equationname{EQ.}

\renewcommand{\vec}[1]{\mathbf{\boldsymbol{#1}}}

\newcommand\expectation{\mathbb{E}}
\newcommand\expectationOver[1]{\expectation_{#1}}

\newcommand{\argmax}{\operatorname{argmax}}

\newcommand{\setnorm}[1]{\left|#1\right|}

\newcommand\labelspace{C}

\newcommand\numberOfClasses{C}
\newcommand\classRunning{c}

\newcommand\inputsingle{\vec{x}}
\newcommand\labelsingle{c}

\newcommand\loss{\mathcal{L}}
\newcommand\prob{p\!}

\newcommand\datasetAll{\mathfrak{D}}

\newcommand\model{f}

\newcommand\inputAL{\inputsingle'}
\newcommand\inputlabelAL{\labelsingle'}

\newcommand\ModelOutputChange{\Delta\model}

\newcommand\rejectionClass{r}

\newcommand\ymulti{\bar{y}(\inputsingle)}

\newcommand\ymultip{y}

\newcommand\kernelFunction{\mathcal{K}}

\newcommand\pPDE{\prob_{\hspace{0.2em} \text{PDE}}}

\newcolumntype{Y}{>{\centering\arraybackslash}X}
\newcolumntype{L}[1]{>{\raggedright\arraybackslash}p{#1}}
\journalname{KI – Künstliche Intelligenz}
\begin{document}
\title{Active and Incremental Learning with Weak Supervision}


\author{Clemens-Alexander Brust \and Christoph Käding \and Joachim Denzler}


\institute{Clemens-Alexander Brust \at
              Friedrich Schiller University Jena \\
              Tel.: +49-3641946425\\
              \email{clemens-alexander.brust@uni-jena.de}
           \and
           Christoph Käding \at
              Friedrich Schiller University Jena \\
              Tel.: +49-3641946425\\
              \email{christoph.kaeding@uni-jena.de}
          \and
          Joachim Denzler \at
              Friedrich Schiller University Jena \\
              Tel.: +49-364146420\\
              \email{joachim.denzler@uni-jena.de}
}

\date{Received: date / Accepted: date.}

\maketitle

{\noindent\footnotesize\emph{Note: This paper is an extended version of our previous work \cite{Brust18_ALO}, from which certain parts of \cref{sec:relw,sec:tasks,sec:actobjdet,sec:exp} (except novel YOLO-specific methods) were taken verbatim. \Cref{sec:emoc} contains some verbatim parts from our previous work \cite{Kaeding15_ALD}.}}

\begin{abstract}
Large amounts of labeled training data are one of the main contributors to the great success that deep models have achieved in the past. Label acquisition for tasks other than benchmarks can pose a challenge due to requirements of both funding and expertise. By selecting unlabeled examples that are promising in terms of model improvement and only asking for respective labels, active learning can increase the efficiency of the labeling process in terms of time and cost.

In this work, we describe combinations of an incremental learning scheme and methods of active learning. These allow for continuous exploration of newly observed unlabeled data. We describe selection criteria based on model uncertainty as well as expected model output change (E\-MOC). An object detection task is evaluated in a continuous exploration context on the PASCAL VOC dataset. We also validate a weakly supervised system based on active and incremental learning in a real-world biodiversity application where images from camera traps are analyzed. Labeling only 32 images by accepting or rejecting proposals generated by our method yields an increase in accuracy from 25.4\% to 42.6\%.

\keywords{Active Learning \and Wildlife Surveillance \and Weak Supervision \and Object Detection \and Incremental Learning}
\end{abstract}

\begin{figure*}[tb]
    \centering
    \includegraphics[width=1.0\textwidth]{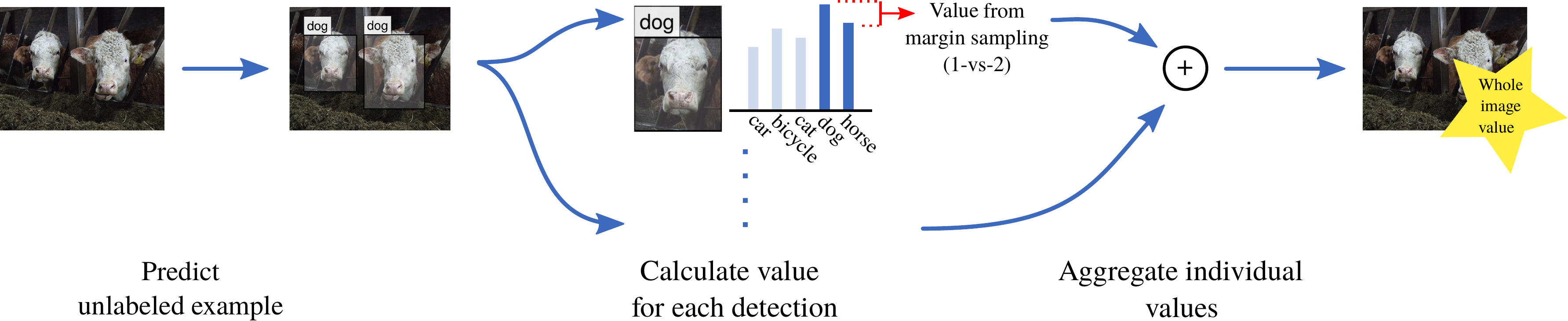}
    \vspace{0pt}
    \caption{
        Pipeline for determining active learning value of a whole unlabeled image in object detection.
        First, bounding boxes are predicted using a method such as YOLO. For each bounding box we calculate an individual
        active learning value, \eg 1-vs-2, based on the distribution of predicted classes. These values are
        then aggregated in order to identify valuable unlabeled images. Figure from \cite{Brust18_ALO}.
        \emph{Note: cows are unknown at this stage in learning.}
    }
    \label{fig:teaser}
\end{figure*}

\section{Introduction}
Deep convolutional networks (CNNs) show impressive performance in a variety of applications.
Even in the challenging task of object detection, they serve as excellent models \cite{Girshick2013RCNN,Liu2015SSD,Long2014FCN,Redmon2015YOLO,Redmon2016YOLOv2}.
Traditionally, most research in the area of object detection builds on models trained once on reliable labeled data for a predefined application.
However, in many application scenarios, new data becomes available over time or the distribution underlying the problem changes.
When this happens, models are usually retrained from scratch
or have to be refined via either fine\hyp{}tuning \cite{hoffman2014lsda,Long2014FCN} or incremental learning \cite{li2016learning,rebuffi2016icarl}.
In any case, a human expert has to assign labels to identify objects and corresponding classes for every unlabeled example.
When domain knowledge is necessary to assign reliable labels, this is the limiting factor in terms of effort or costs.
For example, cancer experts have to manually annotate hundreds of images to provide accurately labeled data \cite{qaiser2017her2,Rodner17_DBF}.

Changing distributions can also pose a problem because constant relabeling is required. Self-driving cars for example
should not be confused by new types of signage or other changes in regulation and environment. To adapt to them,
additional labels need to be supplied.

In the field of biodiversity, there is a strong demand for reliable
and cost-effective methods of estimating diversity indicators such as
animal abundance and site occupancy \cite{mackenzie2004occupancy}. Manual observation in the field can
be replaced by automated camera traps.
However, images from these cameras require expert analysis to be useful
in biodiversity studies. Machine learning can reduce the amount of expert
work required by generalizing from labeled images. Still, when new
species are observed or environmental changes occur, new labels
may be necessary for stable recognition performance.

The goal of active learning is to minimize this labeling effort by selecting only valuable unlabeled examples for annotation by the human domain expert.
Active learning is widely studied in classification tasks, where different measures of uncertainty are the most common choice for selection
\cite{ertekin2007learning,fu2015batch,hoi2006large,Jain09:ALL,Kapoor10:GP,Kaeding2016WALI,tong2001support}.

\subsection{Structure}
This work is split in two parts:
The first part starts by describing the tasks tackled and the active learning problem in general (see \cref{sec:tasks}).
We propose two active learning methods in \cref{sec:actobjdet}. The first method is uniquely generic and can be applied to any object
detection system unlike most related methods. The second is geared towards one of the most famous deep object detectors: ``You Only Look Once'' -- YOLO \cite{Redmon2015YOLO}.
We add an incremental learning scheme \cite{Kaeding2016FDN} to build an object detection system suitable for active and continuous exploration applications.
We first show the validity and performance of our system in an experiment on a popular benchmark dataset in \cref{sec:exp}.
It is then applied and evaluated in a real-life situation in \cref{sec:ctrap}, helping biodiversity researchers with their analysis of wildlife camera footage.
The software is described in detail in \cref{sec:cd} and available upon request.

In the second part, \cref{sec:emoc} gives an outlook into a more theoretically sound active learning method called Expected Model Output Change (EMOC, \cite{Freytag2014EMOC}). As a first step towards
using it in our application, we validate its performance in a related scenario where object proposals from an unsupervised
method are classified.

\section{Related Work}
\label{sec:relw}
\subsection{Object Detection using CNNs}
An important contribution to object detection based on deep learning
is R-CNN \cite{Girshick2013RCNN}. It delivers a considerable improvement over
previously published sliding window-based approaches. R-CNN employs selective search \cite{uijlings2013selective},
an unsupervised method
to generate region proposals. A pre-trained CNN performs feature extraction.
Linear SVMs (one per class) are used to score the extracted features and a
threshold is applied to filter the large number of proposed regions.
Fast R-CNN~\cite{Girshick2015FastRCNN} and Faster R-CNN~\cite{Ren2015FasterRCNN} offer further improvements.

Object detection can also be performed in a single step, combining localization and classification.
An example of this is YOLO \cite{Redmon2015YOLO}, short for \enquote{You Only Look Once}. The authors train a CNN end-to-end as opposed to
using it simply for feature extraction. After the single pass prediction, some post-processing is required
to turn the model's representation into a list of bounding boxes. This includes thresholding and non-maximum suppression.
Because YOLO can classify and localize independently, it can localize objects of unknown classes robustly.
This property is important for active exploration scenarios where new classes may appear at any time.

A similar approach to YOLO is SSD \cite{Liu2015SSD}.
It also delivers the detections using a complex output encoding.
A number of improvements make it more accurate and faster than YOLO at the same time.
One improvement is the incorporation of prior knowledge about the distribution of bounding box
aspect ratios. SSD also considers multiple scales during prediction.

In \cite{lin2017focal}, a new loss function for single-pass object detectors is
proposed to counter the effects of imbalanced positive and negative examples which
are typical for detection tasks. The authors also propose an efficient implementation
called RetinaNet which combines the accuracy of two-stage approaches with the speed
of single-stage approaches.

YOLOv2 \cite{Redmon2016YOLOv2} improves upon the
original YOLO by including
aspect ratio priors for bounding boxes and more fine\hyp{}grained
feature maps using pass\hyp{}through layers to increase resolution. The
network is trained on multiple scales by resizing during training.
The input size can be changed arbitrarily since it only contains convolutional
and pooling layers \cite{Long2014FCN}. Further small improvements are proposed
in \cite{redmon2018yolov3} as YOLOv3, such as training at more scales and using a
more accurate and efficient feature extraction network.
We use YOLO in favor of YOLOv2/v3, RetinaNet or SSD because of its
relative simplicity \wrt its output encoding.

\subsection{Active Learning for Object Detection}
\label{sec:relwactobjdet}
The authors of \cite{abramson2006active} propose an active learning system for pedestrian detection in videos taken by a camera mounted on the front of a moving car.
Their detection method is based on \mbox{AdaBoost} while sampling of unlabeled instances is realized by hand-tuned thresholding of detections.
Object detection using generalized Hough transform in combination with randomized decision trees, called Hough forests, is presented in \cite{yao2012interactive}.
Here, costs are estimated for annotations, and instances with highest costs are selected for labeling.
This follows the intuition that those examples are most likely to be difficult and therefore considered most valuable.
An active learning approach for satellite images using sliding windows in combination with an SVM classifier and margin sampling is proposed in \cite{bietti2012active}.
The combination of active learning for object detection with crowd sourcing is presented in \cite{vijayanarasimhan2014large}.
A part-based detector for SVM classifiers in combination with hashing is proposed for use in large-scale settings.
Active learning is realized by selecting the most uncertain instances for labeling.
In \cite{roy2016active}, object detection is interpreted as a structured prediction problem using a version space approach in the so called ``difference of features'' space.
The authors propose different margin sampling approaches estimating the future margin of an SVM classifier.

Like our proposed approach, most related methods presented above rely on uncertainty information like least confidence or 1\hyp{}vs\hyp{}2.
However, they are designed for a specific type of object detection and therefore can not be applied directly to the output of YOLO.
Additionally, our method does not propose single objects to the human annotator. It presents whole images and takes labels for every object in the image as input.
We also attempt to exploit information specific to YOLO in a secondary approach and compare it to our proposed generic methods.

\subsection{Active Learning for Deep Architectures}
\label{sec:relwactdel}
In \cite{wang2014new} and \cite{Wang2016CEAL},
uncertainty-based active learning criteria for deep models are proposed.
The authors offer several metrics to estimate model uncertainty, including least confidence, margin or entropy sampling.
Wang \etal additionally describe a self-taught learning scheme,
where the model's prediction is used as a label for further training if uncertainty is below a threshold.
Another type of margin sampling is presented in \cite{stark2015captcha}.
The authors propose querying examples according to the quotient of the highest and second-highest class probability.

The visual detection of defects using a ResNet is presented in \cite{feng2017dal}.
The authors propose two methods: uncertainty sampling (\ie defect probability of {0.5}) and positive sampling (\ie selecting every positive example since they are very rare)
for querying unlabeled instances as model update after labeling.
Another work which presents uncertainty sampling is \cite{liu2017active}.
In addition, a query by committee strategy as well as active learning involving weighted incremental dictionary learning for active learning are proposed.

In the work of \cite{gal2017deep}, several uncertainty-related measures for active learning are proposed.
Since they use Bayesian CNNs, they can make use of the probabilistic output and employ methods like variance sampling, entropy sampling or maximizing mutual information.

All of the works introduced above are tailored to active learning in classification scenarios.
Most of them rely on model uncertainty, similar to our proposed selection criteria.
We are evaluating object detection scenarios where we partly take advantage of the special output generated by YOLO.
Thus, these works can not be applied directly.

Besides estimating the uncertainty of the model, further retraining-based approaches are maximizing the expected model change \cite{huang2016active}
or the expected model output change \cite{Kaeding2016AEMOC} that unlabeled examples would cause after labeling.
Since each bounding box inside an image has to be evaluated according its active learning value,
both measures would be impractical in terms of runtime without further modifications.
First steps towards using EMOC for detection are outlined in \cref{sec:emoc}.

A more complete overview of general active learning strategies can be found in \cite{Grauman2016Crowd,Settles2009ALS}.

\subsection{Human-Computer Interaction}
While efficient sampling of unlabeled data for later annotation is an important step towards
better use of expert time and funding, there are other aspects of a learning system that have
potential for improvement, namely the human-computer interaction. Weak supervision in general is the use of labels or supervision signals
that are less precise, accurate or complex than actually required by the task at hand \cite{Zhou2017}. This usually means that more
labels are needed to reach a certain accuracy. However, interaction times can be faster, leading to a
net gain for certain weakly supervised methods. Less precise labels may also be more widely available or cheaper.

In \cite{Papdopoulos2016Semi}, the authors propose an interaction scheme for annotating bounding boxes.
Proposals for bounding boxes are generated and the annotator can only verify, or in certain setups, modify them.
This reduces the annotation time substantially compared to manual painting of bounding boxes, but also leads to
verification of bounding boxes that are not perfect. In their experiments, they show that the trade-off works in
favor of their proposed method, the concept of which we also adopt in our system (see \cref{sec:ctrap,sec:cd}).

Extreme clicking \cite{papadopoulos2017extreme} is an approach which requires manual annotation,
but in a more reasonable manner. Instead of requesting the often non-existent top left and bottom
right corners of an object, the user selects four extreme points in the top, bottom, left, and right directions.
This leads to a 5x speedup in interactions without any loss of accuracy.

\subsection{Automated Wildlife Surveillance}
The work \cite{gomez2016animal} presents a study of the effectiveness of different deep learning architectures on deciding first
if an image shows a bird or mammal and deciding the correct mammal set afterwards using the Snapshot Serengeti dataset \cite{swanson2015snapshot}.
Forwarding images with low confidence decisions to a human expert allows for reaching high accuracies.

A related approach is proposed in \cite{norouzzadeh2017automatically} where animals are classified after deciding if an image contains an animal at all.
This work presents a study of different CNN architectures also using the Snapshot Serengeti dataset.
Another study with a deeper evaluation on different subsets of this dataset involving species-level accuracies was presented in \cite{gomez2017towards}.

Animal segmentation using Multi-Layer Robust Principal Component Analysis involving color and texture features was proposed in \cite{giraldo2017camera}.
This approach was further combined with deep learning methods in \cite{giraldo2017automatic}.
Both works are evaluated on camera trap data from a Colombian forest.

In contrast to those approaches, we do not rely on a fixed training set but explicitly acquire new training data to improve our model.
Additionally, we are able to handle images showing more than one animal since we use detection methods instead of assigning whole image labels.

An animal re-identification approach based on object proposals, which are then used to extract faces for classification, is presented in \cite{Brust_2017_ICCV}.
This method also relies on YOLO to generate class-independent proposals.

\section{Background: Classification, Detection, Supervision and the Active Learning Problem}
\label{sec:tasks}
This section serves to introduce the notation used and problems tackled throughout the first part of this work.

\paragraph{Classification} is a machine learning task in which an example $x$, \eg an image or text, from a data space $\mathfrak{D}$ is assigned a class $c$
from a set $C$ of many possible classes, \eg \texttt{cat} or \texttt{dog}. For our purposes, we require a classifier
to not only assign a class $c$, but predict a distribution over all classes $C$ given an example $x$. As such, we define
a classifier function $f_c$ (also called \emph{score}) per class $c \in C$:
\begin{equation}
  f_c: x \mapsto \hat{p}(c|x) \enspace \text{with } \sum_{c\in C} f_c(x) = 1 \enspace .
\end{equation}
In the following sections, we will mostly look at the classifier output in the form of the estimated distribution $\hat{p}$.

\paragraph{Detection} or object detection is a more complex task where a non-fixed number of instances of classes in an image $x$ is both localized and classified.
For a given image $x$, a detector produces $D$ different detections depending on the content of $x$. For each detection (indexed $i$), a
bounding box $B_i=(x, y, w, h)_i$ and class distribution $\hat{p}_i(c,x)$ are estimated:
\begin{equation}
  f_{c,i}: x \mapsto (B_i, \hat{p}_i(c|x)) \enspace \text{with } \sum_{c\in C} f_{c,i}(x) = 1 \enspace .
\end{equation}
The following sections will focus on the estimated distributions $\hat{p}_i$ of a detector.

\subsection{Incremental Learning}
Typically, classifiers and detectors are trained once, \enquote{seeing} all training data, and then used indefinitely without any further adjustments.
For long-running applications, this setup can become problematic: over time, a problem domain can change or extend, \eg to new classes.
Instead of time- and resource-intensive retraining, one can also apply \emph{incremental learning}. Here, an existing model is augmented
such that it learns any new training data without \enquote{forgetting} about the previous observations.

\subsection{Weakly Supervised Learning}
In most cases, classifiers and detectors are trained in a \textit{supervised} fashion, meaning that the training data is made up of
pairs of examples and labels $(x,y)$. In contrast, \textit{unsupervised} learning considers only the examples themselves, without labels.
Examples include clustering methods as well as generative models. \textit{Weakly supervised} learning is a compromise: labels are available,
but are of reduced quality or information content. This technique can be used to trade-off annotation time against label quality in an effort
to achieve better accuracy within a given amount of annotation time.

\subsection{Active Learning}
\textit{Active learning} is the problem of selecting examples $x$ from an \emph{unlabeled} pool $\mathfrak{U}$ for labeling, \eg
by a human annotator, such that the performance of a future machine learning task is maximized when the selected and annotated examples are learned.
The ultimate goal is to increase data efficiency and to minimize the need for manual annotation.
The active learning problem can be rephrased as a \emph{value} assignment,
where higher values indicate
better future performance when the example is labeled and used for training.
Each example $x$ is assigned a value in $[0, 1]$ by a function $v(x)$, also called an active learning \emph{metric}. Selection is then
performed by sorting all candidate unlabeled examples by their value and choosing the desired amount of top examples.

Active learning is often used in conjunction with incremental learning of small batches. Many active learning methods
incorporate the prediction of an existing model into their value function, which might change substantially after learning only a few examples.
As such, a tight feedback loop is important for good data efficiency.

The predictions of a model on unseen, unlabeled examples, can be analyzed for \emph{uncertainty}.
Uncertainty is one of the most common concepts in active learning
\cite{ertekin2007learning,fu2015batch,hoi2006large,Jain09:ALL,Kapoor10:GP,Kaeding2016WALI,tong2001support},
as it serves as a reasonable indicator of valuable examples.

\paragraph{1-vs-2} An estimated distribution $\hat{p}(c|x)$ can be analyzed for indications of uncertainty.
For example, if the difference between the two highest class probabilities is very low, the example
may be located close to a decision boundary. In this case, it can be used to
refine the decision boundary and is therefore valuable. Its value is determined
using the highest scoring classes $c_1$ and $c_2$, and the following definition:
\begin{equation}
 v_{1vs2}(x)~=~1- (\underset{c_1 \in C}{\textrm{max}}\,\hat{p}(c_1|x) -
\underset{c_2 \in C\setminus c_1}{\textrm{max}}\,\hat{p}(c_2|x))\enspace.
\end{equation}

This metric is known as \emph{1\hyp{}vs\hyp{}2} or \emph{margin sampling} \cite{Settles2009ALS}.
We use 1\hyp{}vs\hyp{}2 as part of our methods since its operation is intuitive and it can produce better estimates than
\eg least confidence approaches \cite{Kaeding2016AEMOC}. A possible alternative is outlined in \cref{sec:emoc}.

\section{Our Methods: Active Learning for Deep Object Detection}
\label{sec:actobjdet}
The active learning problem can also be posed for detection tasks. We consider
the value of labeling whole images $x$ even for detection,
as opposed to individual objects or regions.

In this section, we propose two approaches.
First, a method to adapt any
distribution-based active learning metric
for classification to object detection using an aggregation process. This method
is applicable to any object detector whose output contains class scores for each
detected object.
Second, two metrics specific to the YOLO \cite{Redmon2015YOLO}
object detector are described, using implementation-specific information not available to all object
detectors.

\subsection{Aggregated Detection Metrics}
\label{sec:general}
Using a classification metric on a single detection is straightforward, if class
probabilities are available.
However, aggregating metrics for a complete image can be done in many different ways.
Possible aggregations include calculating the sum, the average or the maximum
over all detection values. However, for some aggregations, it is not clear how an image without
any detections should be handled.

\paragraph{Sum}
A straightforward method of aggregation is the sum.
Intuitively, this method prefers images with lots of uncertain detections in them.
When aggregating detections using a sum, empty examples should be valued zero.
It is the neutral element of addition, making it a reasonable value for an empty
sum. A low valuation effectively
delays the selection of empty examples until there are either no better examples
left or the model has improved enough to actually produce detections on them.
It should be noted that the range of this function is not necessarily $[0, 1]$.
The value of a single example $x$ can be calculated from the detections
$D$ in the following way, where $v_{1vs2}(x_i)$ denotes an application of $v_{1vs2}$
\wrt $\hat{p}_i$:
\begin{align}
\notag
v_{Sum}(x)~=~\sum_{i \in D} v_{1vs2}(x_i)\enspace.
\end{align}

\paragraph{Average}
Another possibility is averaging all detection values.
The average is not sensitive to the number of detections, which may make values
more comparable between images.
If an example does not contain any detections, it will be assigned a zero values.  This is an arbitrary
rule because there is no true neutral element \wrt averages.
However, we choose zero to keep the behavior
in line with the other metrics:
\begin{align}
\notag
v_{Avg}(x)~=~ \frac{1}{|D|}\sum_{i \in D} v_{1vs2}(x_i)\enspace.
\end{align}

\paragraph{Maximum}
Finally, individual detection values can be aggregated by calculating
the maximum. This can result in a substantial information loss. However,
it may also prove beneficial because of increased robustness to noise
from many detections.
For the maximum aggregation, a zero value for empty examples is valid.
The maximum is not affected by zero valued detections,
because no single detection's value can be lower than zero:
\begin{align}
\notag
v_{Max}(x)~=~ \underset{i \in D}{\textrm{max}}\enspace v_{1vs2}(x_i)\enspace.
\end{align}

\subsection{YOLO Specific Metrics}
\label{sec:yolospec}
YOLO \cite{Redmon2015YOLO} offers an end\hyp{}to\hyp{}end approach to deep learning\hyp{}based
object detection. Both its high recognition rate and its real\hyp{}time property
are a result of the compact output encoding. A fixed size vector stores
(within certain boundaries) an arbitrary amount of detections.
To achieve this, the image is divided into $S_h \cdot S_v$ equally sized grid cells.
For each cell $i$, class scores $\hat{p}_i(c)$ are predicted. Furthermore,
the model predicts $B$ bounding boxes, including coordinates relative to the
cell's center, dimensions and an estimated confidence value $\hat{C}$ to
describe a region's \enquote{objectness}.
Adapting classification metrics to object detection can be done by evaluating single
detections and aggregating the results for a complete example (\cref{fig:teaser}).
However, the
YOLO detector's output contains information beyond the detections
themselves. Specifically, it predicts a detection confidence $\hat{C}$ between 0 and 1
separately from the classification scores.

When incorporating the model's detection confidence in\-to a metric, the
following detection\hyp{}specific scenarios can be reacted to:
\emph{(i)} An image cell has a very high class score, indicating a confident
classification, but low predicted detection confidence. This situation can
be caused by a missed detection of a known object class.
\emph{(ii)} A cell has very low class scores overall, but a high confidence estimate.
This may indicate an unknown object class.

Note that because of the way YOLO is implemented and the metrics are designed,
the values $v(x)$ are not bound to be in the range $[0,1]$.

\paragraph{Detection\hyp{}Classification Difference}
Either scenario can be considered a valuable example because it represents
uncertainty in the model. We propose the \emph{Detection\hyp{}Classification Difference}
metric. It aims to detect both scenarios by calculating the absolute difference between
the predicted confidence $\hat{C}$ and the highest class score $\hat{p}$:
\begin{align}
\notag
v_{DetClassDiff}(x)~=~\sum_{i=0}^{S_h S_v} \|&\underset{j=1,\dots,B}{\textrm{max}}\,\hat{C}_{i,j}\\&- \underset{c \in C}{\textrm{max}}\,\hat{p}_i(c|x) \|^2\enspace.
\label{incactmeasurediff}
\end{align}

\paragraph{Weighted Cell Sum}
An adapted classification metric can also be enhanced by using additional
information from YOLO, specifically the predicted confidence $\hat{C}$. The adapted metric
is calculated individually for all cells and then aggregated as a weighted sum,
using the predicted confidences $\hat{C}$ for each cell as weights. We adapt the
1\hyp{}vs\hyp{}2 metric similar to the methods from \cref{sec:general},
resulting in the \emph{Weighted Cell Sum} metric:
\begin{align}
\notag
v_{WCellSum}(x)~=~ \sum_{i=0}^{S_h S_v} \|&\underset{j=1,\dots,B}{\textrm{max}}\,\hat{C}_{i,j}
\\&\cdot v_{1vs2}(x_i) \|^2\enspace.
\label{incactmeasure1vs2}
\end{align}

Assuming high confidence estimates $\hat{C}$ (\ie non\hyp{}objects close to zero and
detections close to one), this metric is very similar to the proposed
\emph{Sum} aggregation that operates only on the detections. With perfect
confidence values, the only differences would be the result of post\hyp{}processing,
\eg non\hyp{}maximum suppression \cite{Redmon2015YOLO}, which is rarely necessary.

Note that the average or maximum operations are not applicable here. A weighted average
would produce identical results to the sum as the number of \enquote{detections}, or grid cells, is constant.
A maximum could either ignore the weights, which would likely result in a constant high value,
or take them into account, in which case it approximates the \emph{Max} variant from the previous section.

\begin{algorithm*}
\caption{{\small Detailed description of the experimental protocol. Please note that in an actual continuous learning scenario, new examples are
always added to $\mathfrak{U}$. The loop is never left because $\mathfrak{U}$ is never exhausted. The described splitting process would have
to be applied regularly.}}
\label{alg:main}
\begin{algorithmic}
\Require{Known labeled examples $\mathfrak{L}$, unknown examples $\mathfrak{U}$, initial model $f_0$, active learning metric $v$}
\State
\State $\mathfrak{U} = \mathfrak{U}_1,\mathfrak{U}_2, \ldots \gets$ split of $\mathfrak{U}$ into random batches
\State $f \gets f_0$
\State
\While{$\mathfrak{U}$ is not empty}
  \State calculate predictions for all unlabeled batches in $\mathfrak{U}$ using $f$
  \State $\mathfrak{U}_{best} \gets$ highest scoring unlabeled batch in $\mathfrak{U}$ according to $v$
  \State
  \State $\mathcal{Y}_{best} \gets$ annotations for $\mathfrak{U}_{best}$ \emph{human-machine interaction}
  \State $f \gets$ incrementally train $f$ using $\mathfrak{L}$ and update batch $(\mathfrak{U}_{best}, \mathcal{Y}_{best})$
  \State
  \State $\mathfrak{U} \gets \mathfrak{U}\backslash\mathfrak{U}_{best}$
  \State $\mathfrak{L} \gets \mathfrak{L} \cup (\mathfrak{U}_{best}, \mathcal{Y}_{best})$
\EndWhile
\end{algorithmic}
\end{algorithm*}

\section{Experiment: PASCAL VOC 2012}
\label{sec:exp}
\begin{table*}
  \small
  \centering
    \begin{tabularx}{0.99\textwidth}{l | Y | Y | Y | Y | Y | Y }
\hline
    & 50 Samples& 100 Samples& 150 Samples& 200 Samples& 250 Samples& All Samples\\
    & mAP / AUC & mAP / AUC & mAP / AUC & mAP / AUC & mAP / AUC & mAP / AUC \\
    \hline
    \multicolumn{7}{l}{Baseline}\\
    \hline
    \hspace{2mm}\emph{Random} & $8.7$/$4.3$ & $12.4$/$14.9$ & $15.5$/$28.8$ & $18.7$/$45.9$ & $21.9$/$66.2$ & $32.4$/$264.0$\\
    \hline
    \multicolumn{7}{l}{Our Methods (YOLO Specific)}\\
    \hline
    \hspace{2mm}\textit{Det.-Class. Diff.} & $8.5$/$4.3$ & $12.1$/$14.6$ & $15.5$/$28.4$ & $18.7$/$45.5$ & $21.0$/$65.3$ & $\bf33.3$/$255.3$\\
    \hspace{2mm}\textit{Weighted Cell Sum} & $\bf9.6$/$\bf4.8$ & $12.9$/$\bf16.1$ & $16.6$/$30.8$ & $\bf20.5$/$49.4$ & $21.9$/$70.6$ & $32.2$/$268.1$\\
    \hline
    \multicolumn{7}{l}{Our Methods}\\
    \hline
    \hspace{2mm}\emph{Max} & $9.2$/$4.6$ & $12.9$/$15.7$ & $15.7$/$30.0$ & $19.8$/$47.8$ & $22.6$/$69.0$ & $32.0$/$\bf269.3$\\
    \hspace{2mm}\emph{Avg} & $9.0$/$4.5$ & $12.4$/$15.2$ & $15.8$/$29.2$ & $19.3$/$46.8$ & $\bf22.7$/$67.8$ & $33.3$/$266.4$\\
    \hspace{2mm}\emph{Sum} & $8.5$/$4.2$ & $\bf14.3$/$15.6$ & $\bf17.3$/$\bf31.4$ & $19.8$/$\bf49.9$ & $22.7$/$\bf71.2$ & $32.4$/$268.2$\\
    \hline
  \end{tabularx}

  \vspace{6pt}
  \caption{Validation results on part B of the PASCAL VOC 2012 dataset, \ie new classes only. \textbf{Bold} indicates best results.}
  \label{tbl:table}
\end{table*}

Our goal is to design an application suitable for automated wildlife surveillance based on
camera trap image analysis involving minimal human supervision while ongoing streams of unlabeled input data occur.
However, we cannot evaluate all methods on the camera trap data because of the limited
availability of labels. Therefore, use the PASCAL VOC 2012
dataset~\cite{Everingham2010VOC} to pose two research questions: (i) can any of our proposed metrics perform better than
random selection and (ii) which metric performs best.

We then use the best performer for our camera trap experiment in the next section.

\paragraph{Methods and Baseline}
The methods compared in this experiment are those proposed in the previous section.
First, the 1\hyp{}vs\hyp{}2 metric aggregated using \emph{Sum}, \emph{Max} and \emph{Avg}. Second, the YOLO\hyp{}specific \textit{Detection-Classification Difference}
and \textit{Weighted Cell Sum}.

We use random selection for comparison. To the best of our knowledge, there are no competing active learning methods that value examples for object detection on an image level at this time.

\paragraph{Data}
We use the PASCAL VOC dataset~\cite{Everingham2010VOC} to assess the effects of our methods on learning.
To specifically measure incremental and active learning performance, both training
and validation set are split into parts A and B in two different random ways to obtain more general experimental results.
Part B is considered \enquote{new} and is comprised of images with specific classes depending on the split%
\footnote{%
\emph{bird}, \emph{cow} and \emph{sheep} (first way) or \emph{tvmonitor}, \emph{cat} and \emph{boat} (second way).}%
. Part A contains all other 17 classes
and is used for initial training. The training set for part B contains 605 and 638 images for the first and second way, respectively.

\paragraph{Active Exploration Protocol}
The experiment follows a typical batchwise incremental and active learning setup \cite{Kaeding2016WALI}.
Before an experimental run, the VOC (part B) datasets are divided randomly into unlabeled batches of 10 examples each.
This fixed assignment decreases the probability of
very similar images being selected for the same unlabeled batch compared to always
selecting the highest valued examples, which would lead to less diverse update batches.
This is valuable while dealing with data streams, \eg from camera traps, or data with low intra-class variance.
The unlabeled batch size is a trade-off between a tight feedback loop (smaller batches)
and computational efficiency (larger batches).

The unlabeled batches are assigned a value using the sum of the
active learning metric over all images in the corresponding unlabeled batch as a meta-aggregation.
Other functions such as average or maximum could be considered,
but are beyond the scope of this paper.

The highest valued unlabeled batch is selected
as an update batch for an incremental training step \cite{Kaeding2016FDN}. The network
is updated using the annotations from the dataset in lieu of a human annotator.
Annotations are not needed for update batch selection. This process
is repeated from the point of unlabeled batch valuation until there are no unlabeled batches
left. The assignment of examples to unlabeled batches is not changed during an experimental run, but
between runs.

\begin{figure*}[t]
  \begin{center}
    \includegraphics[width=0.325\linewidth]{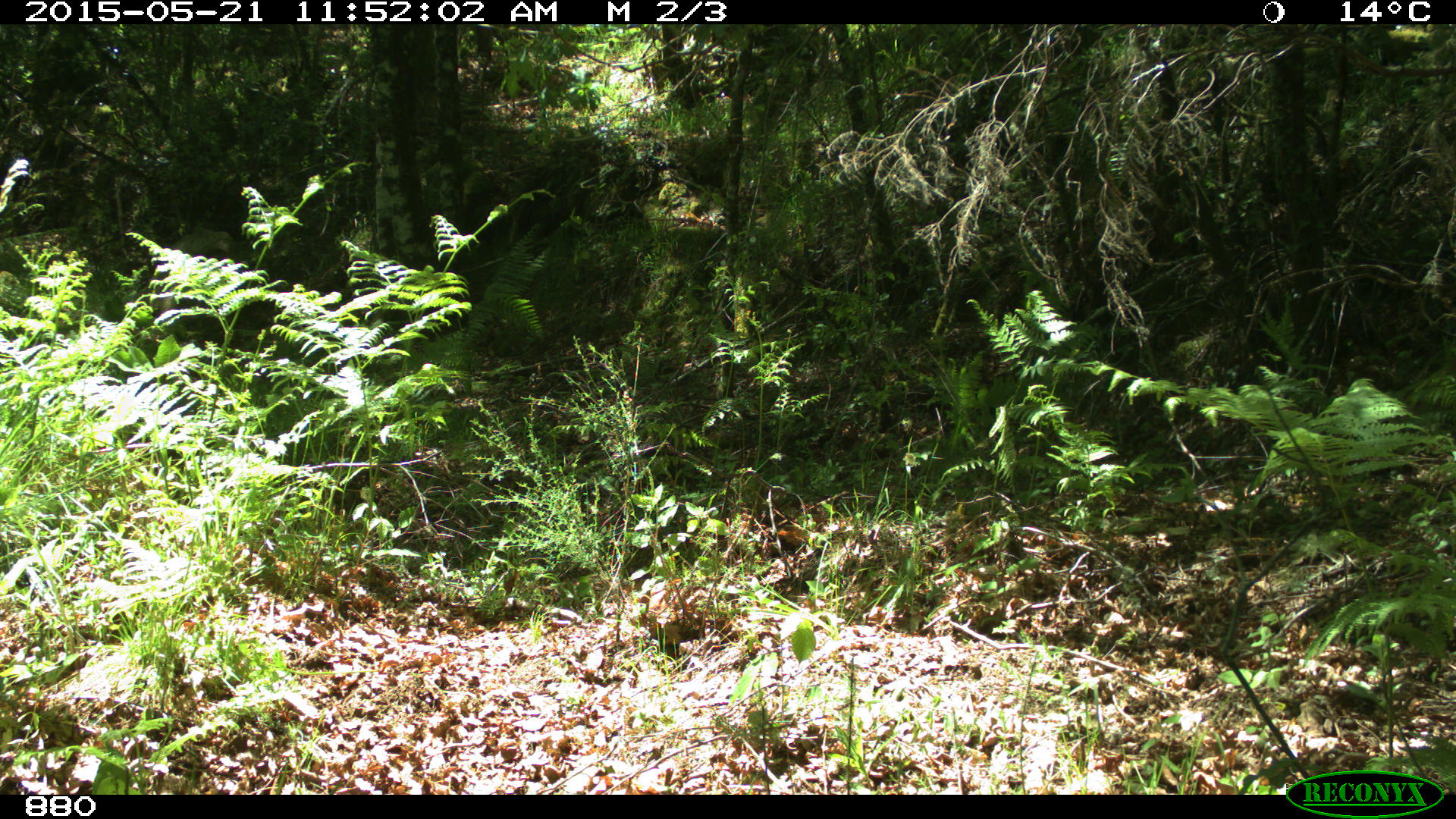}
    \includegraphics[width=0.325\linewidth]{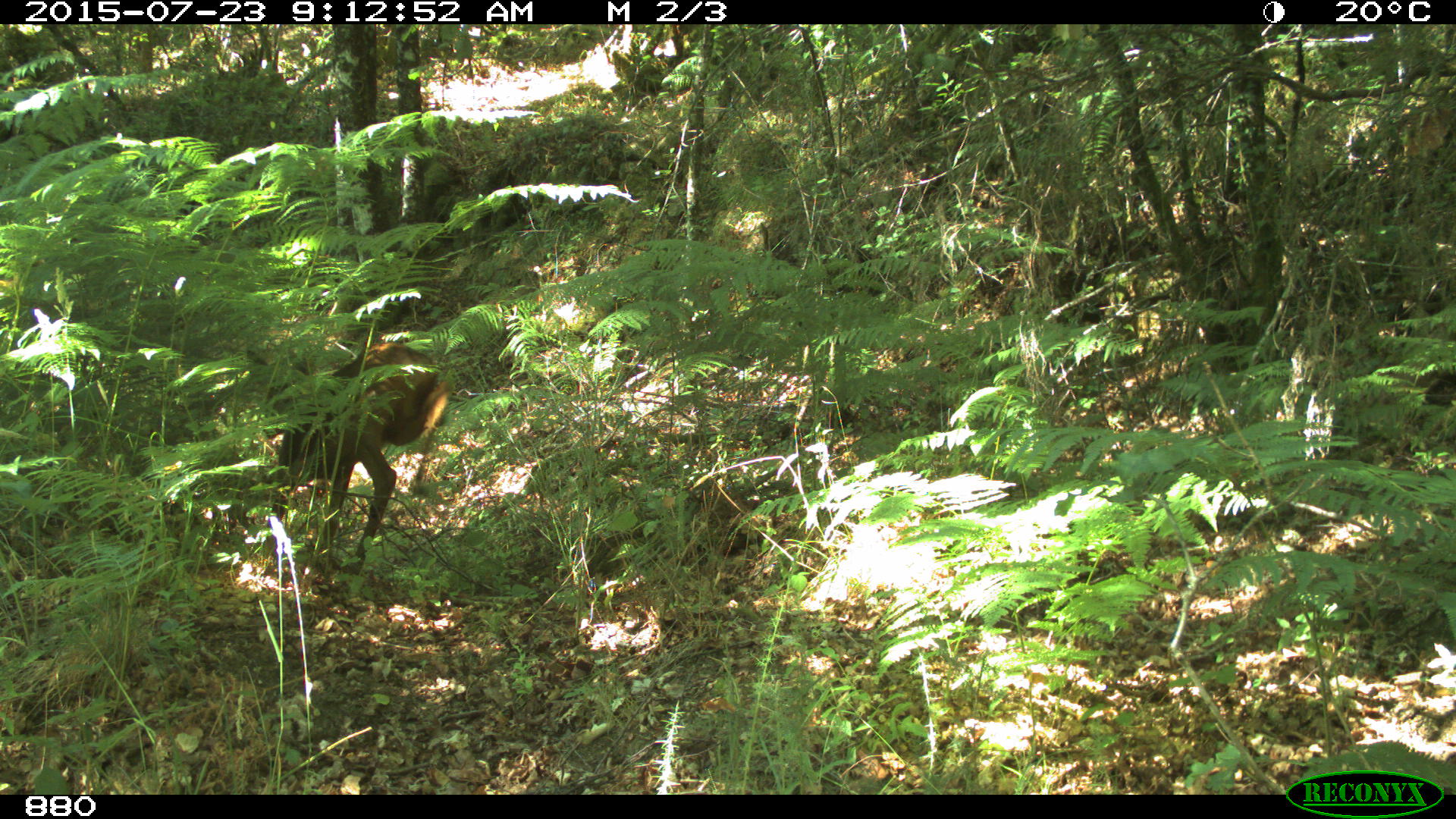}
    \includegraphics[width=0.325\linewidth]{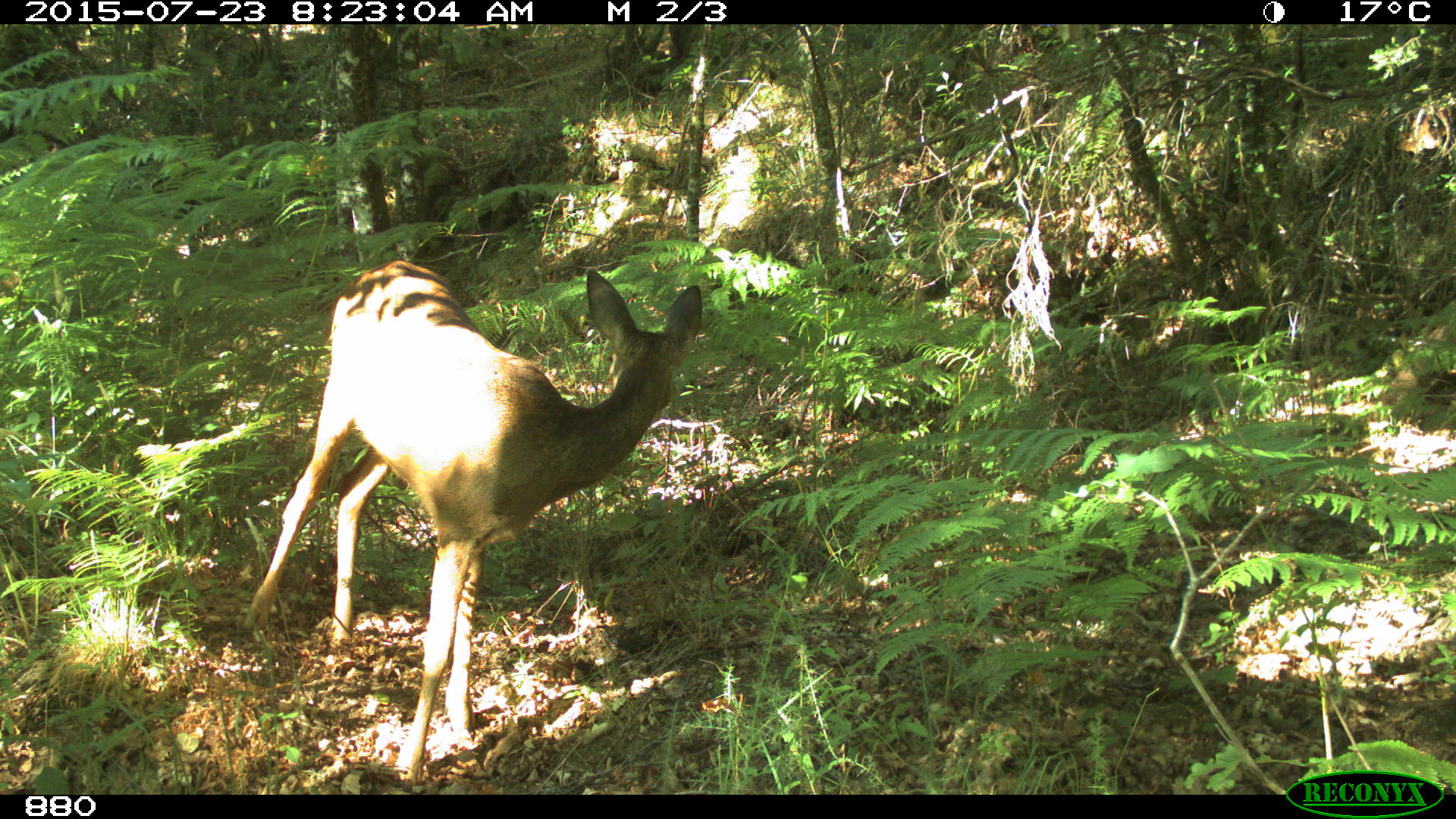}
  \end{center}
  \caption{Camera trap images. From left to right: background only,
  partial occlusion, perfect visibility.}
  \label{fig:bioimgs}
\end{figure*}

\paragraph{Evaluation}
We report mean average precision (mAP) as described in
\cite{Everingham2010VOC}.
For evaluation, we use measures averaged over five runs for each active learning metric as well as random selection, and each
way of splitting.

We show results over the new VOC (part B) classes both in a fast exploration context (\ie after selection of only 150 examples) and after learning all available data.
Gaining accuracy as fast as possible while minimizing the human supervision is one of the main goals of active learning.
Moreover, in continuous exploration scenarios, like faced in camera feeds or other continuous automatic measurements, it is assumed that new data is always available
faster than can be annotated.
Hence, the pool of valuable examples will rarely be exhausted.

We also report AUC, measuring mAP percent points over samples. One unit on the x axis represents 50 samples. The AUC is not normalized and can thus reach a maximum of higher
than 100. It only serves to indicate stability of a method over time and is intended as a sanity check of all methods.

\paragraph{Setup -- YOLO}
We use the YOLO\hyp{}Small architecture
as an alternative to the larger YOLO network, because it allows for much
faster training \cite{Redmon2015YOLO}.
Our initial model is obtained by adapting the \emph{Extraction}
model\footnote{\url{http://pjreddie.com/media/files/extraction.weights}}
and training on the VOC (part A) dataset for 24,000 iterations using the
Adam optimizer~\cite{Kingma2014Adam}. The first half of initial training is completed
with a learning rate of 0.0001. The second half and all incremental experiments
use a lower learning rate of 0.00001 to prevent divergence.
Other hyperparameters match those used
in \cite{Redmon2015YOLO}, including the augmentation of training data using random crops, exposure or saturation adjustments.
The implementation is done in CN24 \cite{Brust2015CPN}, an open-source deep learning framework.

\paragraph{Setup -- Incremental Learning}
Extending an existing CNN without sacrificing performance on known data is not a trivial task.
Fine\hyp{}tuning a CNN exclusively on new data quickly leads to a severe degradation of recognition rates on previously learned examples~\cite{Kirkpatrick2016Cat,Shmelkov2017Detector}.

We use our straightforward, but effective fine\hyp{}tuning me\-thod \cite{Kaeding2016FDN} to implement incremental learning.
With each gradient step, the mini-batch is constructed by randomly selecting from old and new examples with a certain probability of $\lambda$ or $1-\lambda$, respectively.
After completing the learning step, the new data is simply considered old data for the next step.
Management of per\hyp{}example selection probabilities is not necessary.
This method can balance known and unknown data performance successfully.
We use a value of 0.5 for $\lambda$ to make as few assumptions as possible
and perform 100 iterations per update.

Algorithm 1 contains a detailed description of the training procedure.
In our experiments, the cycle ends after all examples are labeled. In a real-world
scenario, the algorithm never leaves the loop because new unlabeled examples
are added continuously.


\subsection{Results}
The learning characteristics of each proposed method on the
new classes from VOC (part B) are shown in \cref{tbl:table}.
In our case, the number of examples added equals the number of images in our experiment.
Validation is performed each time after adding 50 new examples to the current model. We focus our analysis on the new,
unknown data. However, not losing performance on known data is also important.
The incremental learning method from \cite{Kaeding2016FDN}
causes only minimal losses on known data. In the worst case, the mAP
on part A of the VOC dataset decreases from 36.7\% to 31.9\%. These losses are also
referred to as \enquote{catastrophic forgetting} in literature \cite{Kirkpatrick2016Cat}.
The fine-tuning method does not require additional parameters or memory for added examples
like comparable approaches such as \cite{Shmelkov2017Detector} do. This property is an important
step towards \enquote{lifelong learning}, where learning systems can run indefinitely.

\paragraph{Evaluation}
To assess the performance of our methods in a fast exploration context, we evaluate the models after learning 150 examples.
At this point there is still a large number of diverse examples for the methods to choose from,
which makes the following results much more relevant for practical applications than results on the full dataset.

We see \emph{Detection-Classification Difference} perform wor\-st in fast exploration.
Random selection offers comparable results with less variance.
\emph{Average} and \emph{Maximum} perform almost equal to random selection with a very slight advantage.
The best performing fast exploration method is \emph{Sum} with an mAP score of 17.3\%,
improving the random baseline by 1.8\%.
\emph{Weighted Cell Sum} shows similar characteristics with an improvement of 1.1\%.
This result falls in line with our hypothesis that both methods should show similar selection behavior because, under ideal conditions, they perform the same calculations (see \cref{sec:yolospec}).

Surprisingly, metrics specific to YOLO do not generally perform better than the aggregation-based methods.
They may be more sensitive to noise because they are calculated before YOLO's thresholding operation.
Another possible reason is non-maximum suppression. However, it is unlikely
as it only affects a small number of cases~\cite{Redmon2015YOLO}.

\paragraph{All Available Examples}
In our case, active learning only affects the sequence of unlabeled batches if we train until there is no new data available.
Therefore, there are only very small differences between each method's results after training has completed.
However, in continuous exploration, it is usually assumed that there will be more new unlabeled data available than can be processed.
Nevertheless, evaluating the long term performance of our metrics is important to detect possible deterioration over time compared to random selection.
\emph{Detection\hyp{}Classification Difference} achieves the best results by a very small margin when querying all possible training examples.
These small differences also indicate that the chosen incremental learning technique is suitable for the faced scenario.

\paragraph{Discussion}
From the results, we conclude two points: (i) random selection can be outperformed by some of our active learning metrics, and (ii) the \emph{Sum} aggregated
detection metric performs best. After this result, we use the \emph{Sum} metric in the following section.

\section{Experiment: Camera Trap Image Analysis}
\label{sec:ctrap}

\begin{figure}
  \begin{center}
    \includegraphics[width=3.3in]{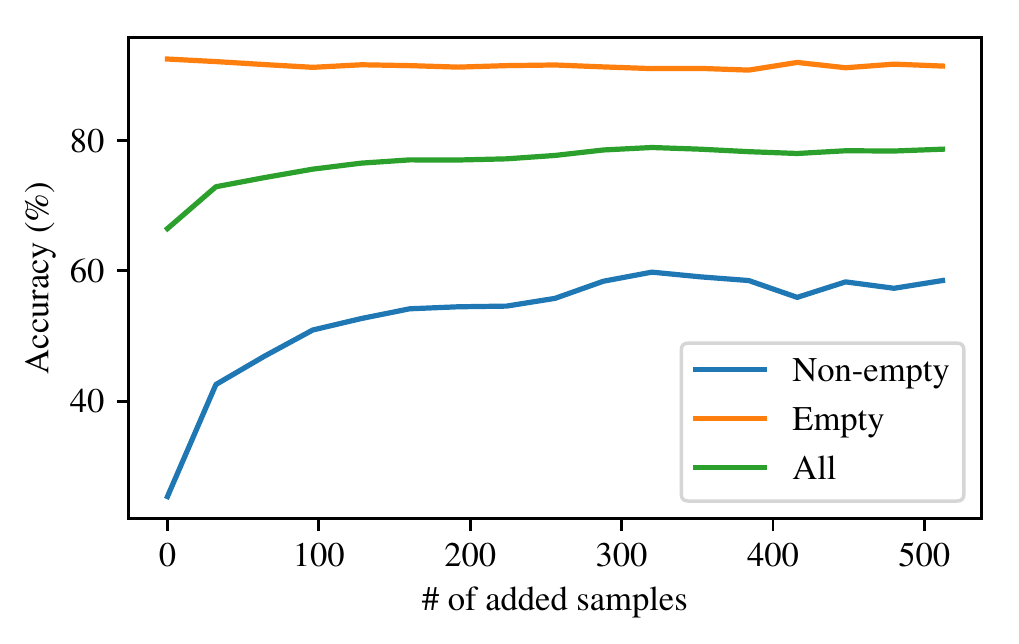}
  \end{center}
  \vspace{-15pt}
  \caption{Camera trap image results from validation set.}
  \label{fig:bioplot}
\end{figure}
After validating the correct operation of our incremental and active learning system
on the PASCAL VOC benchmark dataset, we apply it to camera trap image analysis
in the field of biodiversity. This is to answer an important question:
can the proposed method be applied successfully in real-life scenarios?
For this application, we implement a weakly supervised
system where users
are asked to label images selected using our proposed \emph{Sum} metric, which performed best in the
previous experiment's fast exploration scenario. It tends to favor images with many proposed bounding boxes in it.
Labels are acquired in a propose-and-confirm fashion to increase
efficiency \cite{Papdopoulos2016Semi}. The system is described in detail in \cref{sec:cd}.
The target application is represented by a large biodiversity dataset
created in the course of a project at the German Centre for Integrative Biodiversity Research (iDiv) studying the
impact of large herbivorous mammals on forest development in the National
Park of Peneda-Gerês in Northern Portugal. Up to 65 cameras were deployed
in an area of $\sim$16 km\textsuperscript{2} for a period of 3-4 months in the years 2015
and 2016, resulting in a dataset of around 1.5 million images. The cameras
captured around 15 species of mammals%
\footnote{mainly \emph{cattle, horses, sheep, goats,
wild boar} and \emph{deer}, but also rarer species like \emph{badgers, genets}
and \emph{foxes}}.

\Cref{fig:bioimgs} shows a variety of conditions present in the dataset.
Animals are often occluded by vegetation, camouflaged on purpose to avoid predators or captured from a large distance.
Further difficulties include motion blur, large herds of animals, time of day,
as well as unintentional triggers of the camera trap by humans or moving
leaves.

\subsection{Evaluation}
After validating our method on PASCAL VOC in the previous chapter, we now test it on a separately annotated part of the dataset
consisting of 5,000 examples with image level class labels only.
For labeling and training, there are another 5,000 images to select from.

To evaluate the detector in spite of missing bounding box annotations,
its output is interpreted as a multi-label classifier output.
All other parameters match those detailed in \cref{sec:exp}.
By mapping the classes
of the PASCAL VOC dataset \cite{Everingham2010VOC} to the observed
species, the initial model achieves an accuracy of 66.5\%.
After labeling 512 of the 5,000 training images selected by the \emph{Sum} method
using experts in a fast exploration-like scenario,
the accuracy increases to 78.7\%.

Only 37.8\% of images in the dataset contain objects. \Cref{fig:bioplot}
shows results on empty and non-empty images separately. On the non-empty
subset, accuracy increases from 25.4\% to 42.6\% after labeling only 32 examples,
reaching a final value of 58.5\%.

Longer-term usage could improve the model even further. Weakly supervised learning
on average requires more labels than fully supervised learning to
achieve the same performance. However, it has an overall advantage
due to much shorter labeling times per image \cite{Papdopoulos2016Semi}.

From this experiment, we conclude that our combination of active and incremental learning
can be applied successfully to a real-life camera trap image analysis scenario.


\section{Software: Carpe Diem Annotation Tool}
\label{sec:cd}
\begin{figure}
  \includegraphics[width=\linewidth]{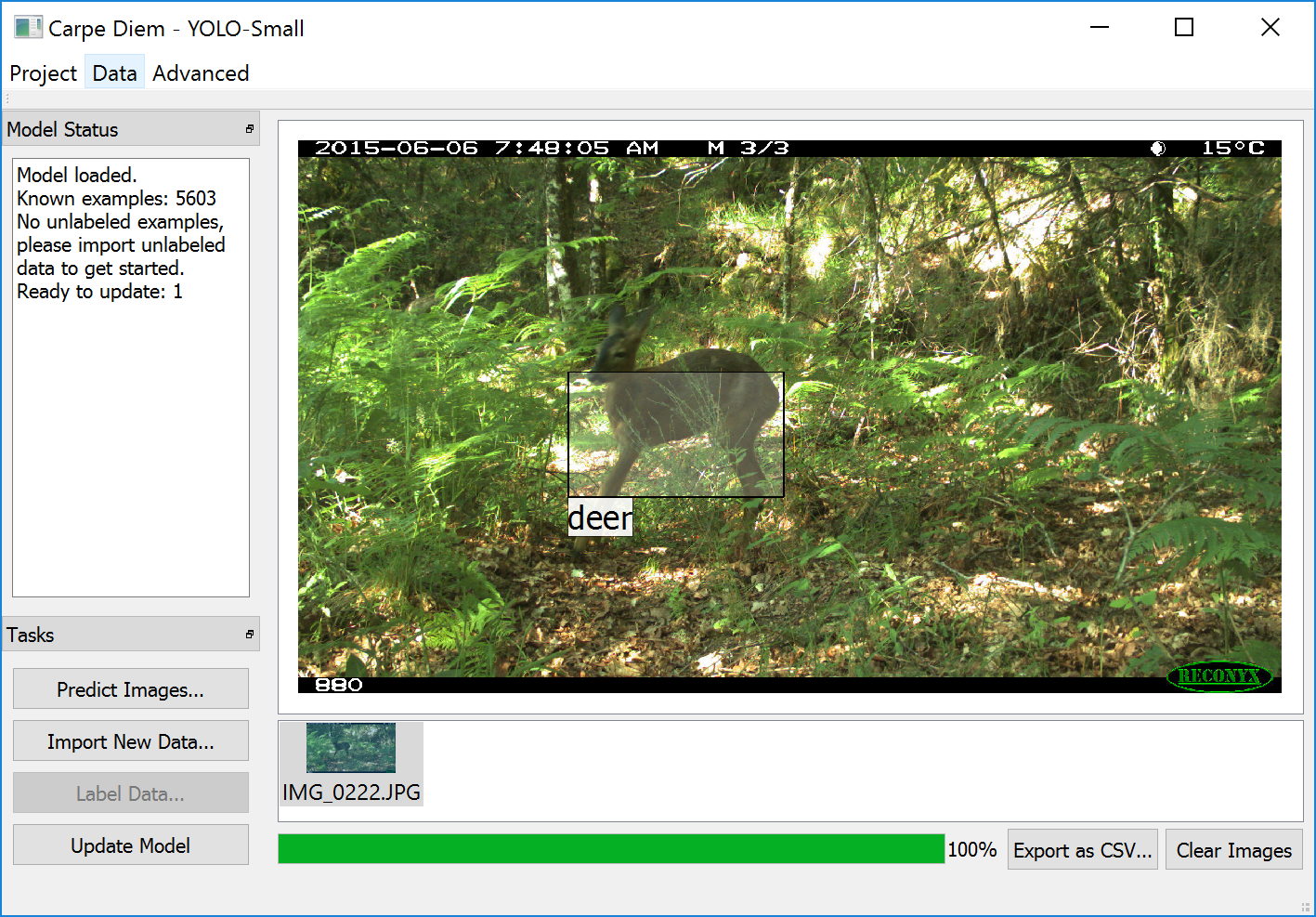}
  \caption{Carpe Diem main screen.}
  \label{fig:cdmain}
\end{figure}

\begin{figure}
  \includegraphics[width=\linewidth]{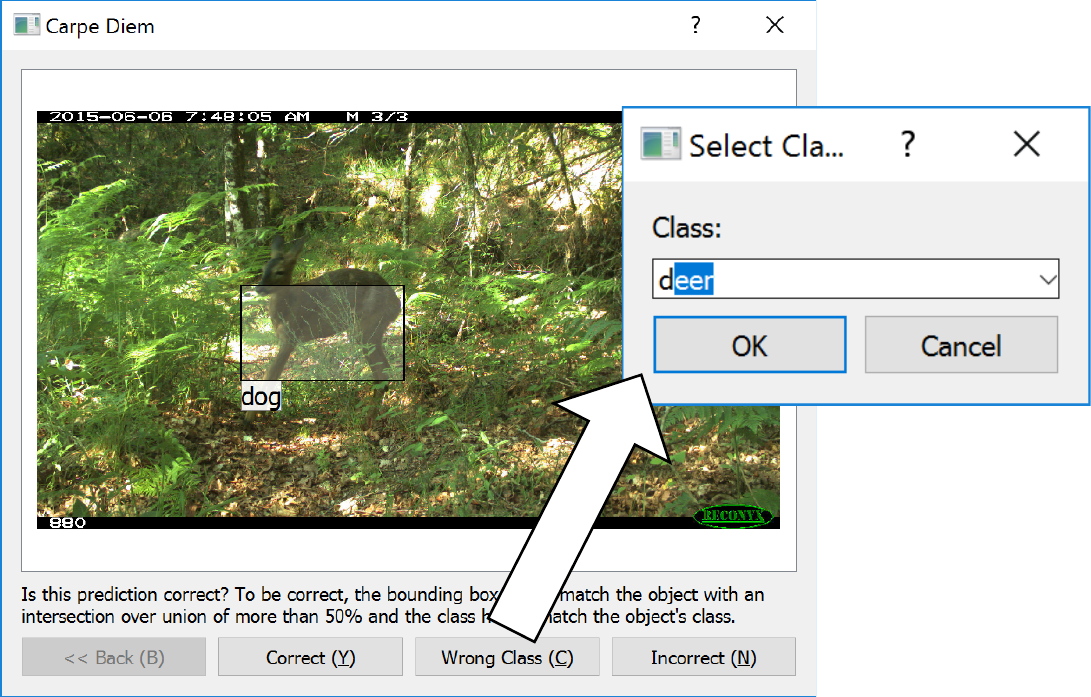}
  \caption{Carpe Diem annotation screen.}
  \label{fig:cdanno}
\end{figure}
In this section, we briefly describe the implementation of our annotation tool offered to biodiversity experts.
This tool, called \emph{Carpe Diem}, realizes a learning cycle environment in a graphical user interface (see \cref{fig:cdmain}).
A learning cycle consists of selection (active learning), label acquisition (user interaction) and model update (incremental learning)
and is executed repeatedly, as new labeling resources become available.
YOLO \cite{Redmon2015YOLO} is used as a detection model and for generating proposals. It is implemented using the CN24 \cite{Brust2015CPN}
framework.

Carpe Diem's clean and simple interface offers all necessary choices and is intuitive to use, even for inexperienced users.
The user can first load or create an annotation project. This collects labeled and unlabeled data as well as a model in one place.
When a project is loaded, the user can generate predictions for images and visualize or export them. Additional labeled
and unlabeled data can be loaded. If there is labeled data that has not yet been observed by the model, a training button
is available.

The main purpose of Carpe Diem lies in the labeling. When the user wishes to label data, a press of a button starts the
evaluation of all unlabeled data against the criteria described in \cref{sec:actobjdet}. The highest scoring batch is then
presented for labeling in a weakly supervised fashion as described in \cref{sec:ctrap}.

The interaction is designed as follows. For each proposed bounding box, the annotator can choose to either confirm it, reject it entirely, or assign a different class (see \cref{fig:cdanno}).
If a reassigned class is unknown, the model will be adapted automatically. After labeling the batch, the user simply
clicks the training button to update the model. The training function takes into account the number of newly labeled images and
uses our incremental learning method presented in \cite{Kaeding2016FDN}.

Carpe Diem is available to researchers on request.

\section{Outlook: Expected Model Output Change}
\label{sec:emoc}

The ultimate goal of active learning is to reduce the risk of models after new examples have been added.
Trying to achieve this in practice reveals substantial problems
such as the absence of the labels necessary to obtain the future risk or the usually small portion of labeled data making it hard to give reliable estimates for the risk
(see \cite{Kaeding15_ALD} for a more detailed introduction on this).

To tackle these problems, the use of surrogates as selection criterion (such as relying on the classification or detection scores like in our methods proposed above) is common practice.
These surrogates show remarkable results in actual applications, such as our wildlife monitoring scenario.
However, researchers also developed approaches using approximations leveraging the search for the smallest future risk.
Some examples for this are \cite{li2014multi,roy2001toward,Settles2009ALS,vijayanarasimhan2011cost,wang2010discriminative}.
In the following, we will briefly review the expected model output change (EMOC) criterion which is indeed an upper bound for the reduction of future risk (a detailed proof is given in \cite{Freytag2014EMOC,Kaeding15_ALD}).
While this approach was already transferred to deep neural networks in \cite{Kaeding2016ACE},
we will demonstrate the performance of the method on an unsupervised detection task using object proposals using Gaussian processes (GP).

This setting is closely related to our application presented in the previous sections. As is, it is not directly applicable to our deep detection scenario,
but could be extended
to the application in the future. This section should thus serve as a self-contained outlook into more theoretically sound methods of active learning, and the
experiments as a first step towards this goal.

Additionally, we will introduce how the EMOC criterion can be extended to handle \emph{unnameable instances}. These are queries that cannot be answered
by the annotator, possibly because of lack of expertise. In our wildlife monitoring application, such cases are to be expected and should be handled
properly.

\subsection{Definition of EMOC}
As introduced, the estimation of risk reduction incurred by a newly labeled example has to deal with severe problems.
To leverage this, \cite{Freytag2014EMOC,Kaeding15_ALD} proposes to favor the selection of unlabeled examples that are most likely to change the model output into any direction.
While this can be traced back to maximizing an upper bound on error reduction from a theoretical perspective, a more intuitive interpretation would be to search for information that \enquote{shake the view on the world} of the current model.
The resulting EMOC criterion can be formalized as follows:
\begin{align}
    \ModelOutputChange(\inputAL)
        &= \expectationOver{\inputlabelAL \in \labelspace}  \; \expectationOver{\inputsingle \in \datasetAll} \left( \loss\left( \model(\inputsingle), \model'(\inputsingle)\right) \right)
    \enspace.
\end{align}
Here, $\model(\cdot)$ stands for the current model while $\model'(\cdot)$ is the future model updated with the new example $(\inputAL,\inputlabelAL)$.
Since the label $\inputlabelAL$ is unknown, the final values is marginalized over every possible known class in the label space $\labelspace$.
We also experimented with explicitly incorporating the possibility of new classes, but found no superior behavior given a more complex estimate.
Furthermore, the change is estimated over the whole available input space $\datasetAll$ which includes known as well as unlabeled examples.
This general formulation still requires to be implemented.
Hence, we will shed some light on a suitable realization in the following.

\paragraph{Choice of the Model Function}
In the following we will rely on GPs which allow for closed form model updates, \ie the step from $\model(\cdot)$ to $\model'(\cdot)$ (see \cite{Freytag13_LET}).
This is beneficial for two reasons.
First, the EMOC criterion can be reformalized which allows for a much more efficient computation.
Second, the actual update after an example is selected and annotated can be done much faster.
Furthermore, choosing GPs as underlying model family allows for further approximations and application scenarios (see \eg \cite{Kaeding2016AEMOC,Kaeding18_ALR}).

\paragraph{Choice of the Loss Function}
In \cite{Freytag2014EMOC}, the choice of the absolute difference of the model outputs was suggested as a suitable loss function.
Since we use the one-vs-all technique, we learn $\numberOfClasses$ binary classifiers $f_\classRunning$ with GP regression when a classification problem with $\numberOfClasses$ classes is given.
Each of the classifiers gives a continuous classification score $\model_\classRunning(\inputsingle) \in \mathbb{R}$, which is used to perform classification decisions according to:
\begin{align}
    \ymulti
        &= \underset{\classRunning=1 \ldots \numberOfClasses}{\argmax} \; \model_\classRunning(\inputsingle)
    \enspace.
    \label{eq:onevsall}
\end{align}
Combining both aspects leads to the following formalization:
\begin{align}
    \loss_{1}\left( \model\left(\inputsingle\right), \model'\left(\inputsingle\right) \right))
        &= \sum\limits_{\classRunning=1}^{\numberOfClasses} | \model_\classRunning(\inputsingle) - \model_\classRunning'(\inputsingle) |
    \enspace.
\end{align}
While the loss can in general be chosen arbitrarily, we stick to the $\loss_{1}$ loss for the shown experiments.
Please see \cite{Kaeding15_ALD} for an evaluation considering more options.

\paragraph{Choice of Multi-class Classification Probabilities}
We compute multi-class probabilities directly derived from uncertainty estimates~\cite{Froehlich13:GSS}.
The underlying idea of the uncertainty technique is that for label regression with GPs,
we do not only have the model prediction $\model_\classRunning(\inputsingle)$ but rather the whole posterior distribution
$\mathcal{N} ( \model_\classRunning(\inputsingle), \sigma^2(\inputsingle) )$ independently for each binary classification problem involved in the one-vs-all problem.
The probability of class $\classRunning$ achieving the maximum score in \equationname~\eqref{eq:onevsall} can therefore be expressed by:
\begin{align}
\prob\left(\ymulti = \classRunning | \inputsingle\right) &=
           \prob\left( \classRunning =  \underset{\classRunning'=1 \ldots \numberOfClasses}{\argmax} \; \model_{\classRunning'}(\inputsingle) \right) \enspace.
\end{align}
To estimate the probabilities, we apply a Monte-Carlo technique and sample $Z$ times from all $\numberOfClasses$
Gaussian distributions $\mathcal{N}( \model_\classRunning(\inputsingle), \sigma^2(\inputsingle) )$ and estimate
the probability of each class:
\begin{align}
\prob\left(\ymultip = \classRunning | \inputsingle\right) = \prob\left(\ymulti = \classRunning | \inputsingle \right) &\approx \frac{Z_\classRunning}{Z} \enspace,
\end{align}
with $Z_\classRunning$ denoting the number of times where the draw from the distribution of class $\classRunning$ was the maximum value.
A large variance $\sigma^2$, \ie a high uncertainty of the estimate, leads
to a nearly uniform distribution $\prob\left(\ymultip=\classRunning\right)$, whereas a zero variance results
in a distribution which is equal to one for the class which corresponds to the highest posterior mean.
An evaluation considering more options is given in \cite{Kaeding15_ALD}.

\subsection{Active Learning with Unnameable Instances}
A very common assumption in active learning is that the oracle (\eg a human annotator) can provide a label for every instance of the set of unlabeled examples.
Especially for tasks that involve a large set of categories, this assumption is not reasonable.
There may be further complications due to occlusions, which are a large problem in wildlife monitoring and can make it impossible to assign a label.
Therefore, we have to deal with cases where the oracle rejects to label the example that the active learning algorithm just selected.
From our experience, there are basically two main scenarios in which a rejection can possibly happen.
Both cases need to be considered during active learning and we present solutions and adaptations of the EMOC principle for each of them in the following.

\paragraph{Dealing with Non-Categorical Rejections}
An unlabeled example may not show a valid object.
Possible reasons are \emph{noise} during image acquisition (\eg sensor noise, motion blur, or JPEG artifacts), segments covering \emph{multiple objects}, \emph{moving vegetation} setting off
a camera trap, or \emph{background} regions.
Hence, the number of images showing no valid objects may be vast.
However, it is unlikely that during dataset acquisition and proposal generation, the same non-object example is obtained several times.
Thus, examples that do not show valid objects are characterized by a low data density
\footnote{A low data density for non-objects is reasonable, \eg sensor noise should happen rarely, or segment proposals should by algorithmic design favor objects over non-objects.}.
In contrast, examples from object categories should cluster since different examples from the same category are likely to be recorded over time.
Therefore, the examples we query should be in a high density region to ensure a high impact on examples nearby.
In contrast, we propose to use the local data density $\prob\left(\inputAL\right)$ obtained with a Parzen estimate:
\begin{equation}
  \pPDE\left( \inputAL \right) \propto \frac{1}{ \setnorm{\datasetAll} } \sum_{\inputsingle_j \in \datasetAll} \kernelFunction\left( \inputsingle_j, \inputAL \right) \enspace,
\end{equation}
where $\kernelFunction$ is a kernel function measuring example similarity.
Combining this with the EMOC criterion leads to:
\begin{align}
  \label{eq:alEMOCpde}
  \begin{split}
  \ModelOutputChange(\inputAL)  = \!
          &\sum_{\inputlabelAL \in \labelspace} \! \prob\left(\inputlabelAL | \model\left(\inputAL\right)\right) \cdot \pPDE\left(\inputAL\right)\\
          &\;\cdot  \Bigl(
          \sum_{\inputsingle_j \in  \datasetAll} \!\!
                   \loss\left( \model(\inputsingle_j),
                       \model'(\inputsingle_j) \right)
            \Bigr) \enspace.
  \end{split}
\end{align}
This is essential in order to focus on examples in high-density regions rather than on less frequent non-categorical examples.

\paragraph{Dealing with Categorical Rejections}
An unlabeled example may be a valid object, but the annotator is \emph{not able} to name it or he decides that it is \emph{not part} of the problem domain, \ie it belongs to \emph{unknown or unrelated categories}, \eg a researcher walking by the camera.
These examples are referred to as ``blind spots'' by \cite{Fang12:IDK} and we model them as one big class $\rejectionClass$.
In particular, $\inputlabelAL = \rejectionClass$ denotes the event when an annotator would reject the example $\inputAL$ and we need to take this into account when computing the EMOC values.
We make use of the fact that we would not get an additional training example in this case. Thus, the classification model would simply not change, \ie $\forall \inputsingle: \model'(\inputsingle) = \model(\inputsingle)$,
which results in zero expected model output change for the case of $\inputlabelAL = \rejectionClass$.
The EMOC value for an example $\inputAL$ under the assumption that there exists a rejection class $\rejectionClass$ is therefore given by:
\begin{align}
  \notag
  \ModelOutputChange^{\rejectionClass}(\inputAL)  &= \expectationOver{\inputlabelAL \in \labelspace \cup \{\rejectionClass\}}  \; \expectationOver{\inputsingle \in \datasetAll}
        \left( \loss\left( \model(\inputsingle), \model'(\inputsingle)\right) \right)\\
  \notag
                                &= \prob\left(\inputlabelAL \neq \rejectionClass | \inputAL\right) \cdot \ModelOutputChange(\inputAL) + \prob\left(\inputlabelAL = \rejectionClass | \inputAL\right) \cdot 0\\
                                &= (1 - \prob\left(\inputlabelAL = \rejectionClass | \inputAL)\right) \cdot \ModelOutputChange(\inputAL)\enspace.
\end{align}
In practice, we estimate the probability $\prob\left(\inputlabelAL = \rejectionClass | \inputAL\right)$
of an example $\inputAL$ being an unnameable instance by using a GP regression classifier learned with previously rejected instances as positive examples and all examples of known classes as negatives.
The classification score predicted by the classifier is transformed into a valid probability value using the probit model~\cite{Freytag2014EMOC}.
As a byproduct, this allows to also model rejections for non-categorical examples.

\subsection{Active Discovery with Object Proposals}
\newlength{\mytikzplotwidth}
\setlength{\mytikzplotwidth}{.42\textwidth}

\newlength{\mytikzplotheight}
\setlength{\mytikzplotheight}{.15\textheight}

\newlength{\mytikzplotColSepLegend}
\setlength{\mytikzplotColSepLegend}{6pt}

\begin{figure*}
    \centering
    \definecolor{mycolor1}{rgb}{0.29804,0.00000,0.60000}%
\definecolor{mycolor2}{rgb}{1.00000,0.60000,1.00000}%
\definecolor{mycolor3}{rgb}{0.00000,0.50000,0.90000}%
\definecolor{mycolor4}{rgb}{1.00000,0.00000,1.00000}%

\newenvironment{customlegend}[1][]{
    \begingroup
    \csname pgfplots@init@cleared@structures\endcsname
    \pgfplotsset{#1}
}{
    \csname pgfplots@createlegend\endcsname
    \endgroup
}

\def\addlegendimage{\csname pgfplots@addlegendimage\endcsname}%

\begin{tikzpicture}

    \tikzstyle{every node}=[font=\scriptsize]

    \begin{customlegend}[
            legend entries={
                Random\\%
                GP-Var~\cite{Kapoor10:GP}\\%
                GP-Unc~\cite{Kapoor10:GP}\\%
                1-vs-2~\cite{Joshi09:MAL}\\%
                PKNN~\cite{Jain09:ALL}\\%
                ERM~\cite{roy2001toward}\\%
                \textbf{GP-EMOC}\\%
            },
            legend cell align=left,
            legend columns=7,
            legend style={
                draw=none,
                /tikz/column 2/.style={column sep=\mytikzplotColSepLegend,},
                /tikz/column 4/.style={column sep=\mytikzplotColSepLegend,},
                /tikz/column 6/.style={column sep=\mytikzplotColSepLegend,},
                /tikz/column 8/.style={column sep=\mytikzplotColSepLegend,},
            },
        ]
    ]
    \addlegendimage{color=black,dashed,line width=1.5pt}
    \addlegendimage{color=mycolor1,dashed,line width=1.5pt}
    \addlegendimage{color=mycolor2,dashed,line width=1.5pt}
    \addlegendimage{color=white!60!green,dashed,line width=1.5pt}
    \addlegendimage{color=gray,dashed,line width=1.5pt}
    \addlegendimage{color=mycolor3,dashed,line width=1.5pt}
    \addlegendimage{color=red,solid,line width=1.5pt}
    \end{customlegend}
\end{tikzpicture}
    \\
    \scriptsize
%
%
\definecolor{mycolor1}{rgb}{0.29804,0.00000,0.60000}%
\definecolor{mycolor2}{rgb}{1.00000,0.60000,1.00000}%
\definecolor{mycolor3}{rgb}{0.00000,0.50000,0.90000}%
\definecolor{mycolor4}{rgb}{1.00000,0.00000,1.00000}%
\begin{tikzpicture}

\begin{axis}[%
width=0.967346\mytikzplotwidth,
height=\mytikzplotheight,
at={(0\mytikzplotwidth,0\mytikzplotheight)},
scale only axis,
xmin=0,
xmax=100,
xlabel={\# Added Samples},
ymin=1.9,
ymax=10.1,
ylabel={\# Discovered Classes},
axis x line*=bottom,
axis y line*=left,
x label style={at={(axis description cs:0.5,0.0)}},
y label style={at={(axis description cs:0.05,0.5)}},
]
\addplot [color=black,dashed,line width=1.5pt,forget plot]
  table[row sep=crcr]{%
0	2\\
1	2.48\\
2	2.91\\
3	3.23\\
4	3.46\\
5	3.77\\
6	4.06\\
7	4.35\\
8	4.6\\
9	4.85\\
10	5.08\\
11	5.24\\
12	5.55\\
13	5.72\\
14	5.95\\
15	6.1\\
16	6.25\\
17	6.44\\
18	6.53\\
19	6.63\\
20	6.8\\
21	6.93\\
22	7.08\\
23	7.21\\
24	7.3\\
25	7.36\\
26	7.46\\
27	7.54\\
28	7.64\\
29	7.7\\
30	7.79\\
31	7.91\\
32	8.04\\
33	8.1\\
34	8.17\\
35	8.28\\
36	8.32\\
37	8.38\\
38	8.51\\
39	8.56\\
40	8.63\\
41	8.68\\
42	8.72\\
43	8.76\\
44	8.81\\
45	8.88\\
46	8.96\\
47	9.01\\
48	9.06\\
49	9.11\\
50	9.14\\
51	9.15\\
52	9.2\\
53	9.24\\
54	9.27\\
55	9.28\\
56	9.29\\
57	9.3\\
58	9.35\\
59	9.38\\
60	9.39\\
61	9.42\\
62	9.42\\
63	9.44\\
64	9.45\\
65	9.45\\
66	9.47\\
67	9.48\\
68	9.49\\
69	9.52\\
70	9.54\\
71	9.56\\
72	9.58\\
73	9.6\\
74	9.61\\
75	9.61\\
76	9.61\\
77	9.61\\
78	9.61\\
79	9.62\\
80	9.63\\
81	9.65\\
82	9.66\\
83	9.68\\
84	9.68\\
85	9.68\\
86	9.69\\
87	9.7\\
88	9.71\\
89	9.72\\
90	9.73\\
91	9.73\\
92	9.73\\
93	9.73\\
94	9.74\\
95	9.74\\
96	9.75\\
97	9.75\\
98	9.75\\
99	9.76\\
100	9.78\\
};
\addplot [color=mycolor1,dashed,line width=1.5pt,forget plot]
  table[row sep=crcr]{%
0	2\\
1	2.71\\
2	3.37\\
3	3.52\\
4	3.74\\
5	4.25\\
6	4.61\\
7	4.97\\
8	5.44\\
9	5.97\\
10	6.19\\
11	6.61\\
12	6.77\\
13	6.84\\
14	7.02\\
15	7.32\\
16	7.48\\
17	7.61\\
18	7.74\\
19	8.08\\
20	8.25\\
21	8.45\\
22	8.65\\
23	8.75\\
24	8.89\\
25	8.99\\
26	9.06\\
27	9.09\\
28	9.1\\
29	9.1\\
30	9.1\\
31	9.1\\
32	9.1\\
33	9.1\\
34	9.1\\
35	9.1\\
36	9.1\\
37	9.1\\
38	9.1\\
39	9.1\\
40	9.1\\
41	9.1\\
42	9.1\\
43	9.1\\
44	9.1\\
45	9.1\\
46	9.1\\
47	9.1\\
48	9.1\\
49	9.1\\
50	9.1\\
51	9.1\\
52	9.1\\
53	9.1\\
54	9.1\\
55	9.1\\
56	9.1\\
57	9.1\\
58	9.1\\
59	9.1\\
60	9.1\\
61	9.1\\
62	9.1\\
63	9.1\\
64	9.1\\
65	9.1\\
66	9.1\\
67	9.1\\
68	9.1\\
69	9.1\\
70	9.1\\
71	9.1\\
72	9.1\\
73	9.1\\
74	9.1\\
75	9.1\\
76	9.1\\
77	9.1\\
78	9.1\\
79	9.1\\
80	9.1\\
81	9.1\\
82	9.1\\
83	9.12\\
84	9.14\\
85	9.15\\
86	9.26\\
87	9.35\\
88	9.53\\
89	9.67\\
90	9.81\\
91	9.9\\
92	9.95\\
93	9.98\\
94	10\\
95	10\\
96	10\\
97	10\\
98	10\\
99	10\\
100	10\\
};
\addplot [color=mycolor2,dashed,line width=1.5pt,forget plot]
  table[row sep=crcr]{%
0	2\\
1	2.5\\
2	3.11\\
3	3.52\\
4	3.93\\
5	4.24\\
6	4.53\\
7	4.81\\
8	5.05\\
9	5.32\\
10	5.58\\
11	5.92\\
12	6.19\\
13	6.43\\
14	6.67\\
15	6.88\\
16	7.09\\
17	7.25\\
18	7.37\\
19	7.48\\
20	7.6\\
21	7.74\\
22	7.87\\
23	7.96\\
24	8.09\\
25	8.18\\
26	8.25\\
27	8.31\\
28	8.39\\
29	8.47\\
30	8.5\\
31	8.55\\
32	8.67\\
33	8.74\\
34	8.8\\
35	8.86\\
36	8.9\\
37	8.95\\
38	8.97\\
39	8.99\\
40	9.02\\
41	9.03\\
42	9.05\\
43	9.06\\
44	9.12\\
45	9.14\\
46	9.16\\
47	9.16\\
48	9.17\\
49	9.18\\
50	9.2\\
51	9.22\\
52	9.22\\
53	9.24\\
54	9.25\\
55	9.27\\
56	9.27\\
57	9.29\\
58	9.29\\
59	9.29\\
60	9.29\\
61	9.32\\
62	9.32\\
63	9.33\\
64	9.34\\
65	9.35\\
66	9.35\\
67	9.37\\
68	9.39\\
69	9.41\\
70	9.42\\
71	9.42\\
72	9.43\\
73	9.43\\
74	9.43\\
75	9.43\\
76	9.44\\
77	9.44\\
78	9.46\\
79	9.47\\
80	9.48\\
81	9.48\\
82	9.48\\
83	9.51\\
84	9.51\\
85	9.51\\
86	9.53\\
87	9.56\\
88	9.57\\
89	9.57\\
90	9.58\\
91	9.59\\
92	9.6\\
93	9.61\\
94	9.64\\
95	9.64\\
96	9.66\\
97	9.68\\
98	9.68\\
99	9.71\\
100	9.72\\
};
\addplot [color=white!60!green,dashed,line width=1.5pt,forget plot]
  table[row sep=crcr]{%
0	2\\
1	2.28\\
2	2.59\\
3	2.99\\
4	3.27\\
5	3.51\\
6	3.75\\
7	4\\
8	4.21\\
9	4.43\\
10	4.68\\
11	4.88\\
12	5\\
13	5.21\\
14	5.39\\
15	5.5\\
16	5.6\\
17	5.67\\
18	5.84\\
19	5.91\\
20	6.02\\
21	6.11\\
22	6.23\\
23	6.31\\
24	6.38\\
25	6.46\\
26	6.53\\
27	6.65\\
28	6.75\\
29	6.81\\
30	6.88\\
31	7.02\\
32	7.11\\
33	7.18\\
34	7.24\\
35	7.33\\
36	7.39\\
37	7.49\\
38	7.58\\
39	7.61\\
40	7.67\\
41	7.73\\
42	7.79\\
43	7.85\\
44	7.93\\
45	7.96\\
46	7.99\\
47	8.06\\
48	8.1\\
49	8.13\\
50	8.16\\
51	8.2\\
52	8.24\\
53	8.26\\
54	8.28\\
55	8.32\\
56	8.34\\
57	8.36\\
58	8.39\\
59	8.41\\
60	8.48\\
61	8.5\\
62	8.54\\
63	8.55\\
64	8.59\\
65	8.61\\
66	8.65\\
67	8.67\\
68	8.69\\
69	8.72\\
70	8.75\\
71	8.78\\
72	8.8\\
73	8.8\\
74	8.86\\
75	8.89\\
76	8.9\\
77	8.94\\
78	8.96\\
79	9\\
80	9.02\\
81	9.03\\
82	9.03\\
83	9.05\\
84	9.07\\
85	9.07\\
86	9.08\\
87	9.1\\
88	9.12\\
89	9.14\\
90	9.17\\
91	9.17\\
92	9.17\\
93	9.18\\
94	9.23\\
95	9.26\\
96	9.26\\
97	9.28\\
98	9.31\\
99	9.32\\
100	9.34\\
};
\addplot [color=gray,dashed,line width=1.5pt,forget plot]
  table[row sep=crcr]{%
0	2\\
1	2.25\\
2	2.47\\
3	2.68\\
4	2.93\\
5	3.14\\
6	3.44\\
7	3.59\\
8	3.79\\
9	4.02\\
10	4.2\\
11	4.35\\
12	4.54\\
13	4.71\\
14	4.92\\
15	5.07\\
16	5.25\\
17	5.43\\
18	5.51\\
19	5.68\\
20	5.76\\
21	5.86\\
22	6.03\\
23	6.14\\
24	6.21\\
25	6.35\\
26	6.47\\
27	6.56\\
28	6.62\\
29	6.7\\
30	6.76\\
31	6.84\\
32	6.91\\
33	6.98\\
34	7.1\\
35	7.2\\
36	7.24\\
37	7.35\\
38	7.39\\
39	7.44\\
40	7.5\\
41	7.53\\
42	7.58\\
43	7.62\\
44	7.7\\
45	7.77\\
46	7.86\\
47	7.9\\
48	7.98\\
49	8\\
50	8.05\\
51	8.07\\
52	8.14\\
53	8.19\\
54	8.23\\
55	8.26\\
56	8.29\\
57	8.31\\
58	8.35\\
59	8.4\\
60	8.42\\
61	8.48\\
62	8.5\\
63	8.56\\
64	8.62\\
65	8.66\\
66	8.7\\
67	8.75\\
68	8.82\\
69	8.88\\
70	8.89\\
71	8.92\\
72	8.94\\
73	8.96\\
74	8.98\\
75	9\\
76	9.03\\
77	9.03\\
78	9.08\\
79	9.12\\
80	9.16\\
81	9.16\\
82	9.19\\
83	9.2\\
84	9.22\\
85	9.22\\
86	9.22\\
87	9.23\\
88	9.24\\
89	9.24\\
90	9.25\\
91	9.26\\
92	9.28\\
93	9.3\\
94	9.32\\
95	9.34\\
96	9.35\\
97	9.37\\
98	9.37\\
99	9.4\\
100	9.4\\
};
\addplot [color=mycolor3,dashed,line width=1.5pt,forget plot]
  table[row sep=crcr]{%
0	2\\
1	2.79\\
2	3.38\\
3	3.72\\
4	4.21\\
5	4.49\\
6	4.97\\
7	5.31\\
8	5.7\\
9	6.01\\
10	6.47\\
11	6.72\\
12	6.79\\
13	6.9\\
14	7.08\\
15	7.29\\
16	7.5\\
17	7.63\\
18	7.7\\
19	7.86\\
20	8.1\\
21	8.35\\
22	8.5\\
23	8.65\\
24	8.78\\
25	8.88\\
26	9.03\\
27	9.08\\
28	9.09\\
29	9.1\\
30	9.1\\
31	9.1\\
32	9.1\\
33	9.1\\
34	9.1\\
35	9.1\\
36	9.1\\
37	9.1\\
38	9.1\\
39	9.1\\
40	9.1\\
41	9.1\\
42	9.1\\
43	9.1\\
44	9.1\\
45	9.1\\
46	9.1\\
47	9.1\\
48	9.1\\
49	9.1\\
50	9.1\\
51	9.1\\
52	9.1\\
53	9.1\\
54	9.1\\
55	9.1\\
56	9.1\\
57	9.1\\
58	9.1\\
59	9.1\\
60	9.1\\
61	9.1\\
62	9.1\\
63	9.1\\
64	9.1\\
65	9.1\\
66	9.1\\
67	9.1\\
68	9.1\\
69	9.1\\
70	9.1\\
71	9.1\\
72	9.1\\
73	9.1\\
74	9.1\\
75	9.1\\
76	9.1\\
77	9.1\\
78	9.1\\
79	9.1\\
80	9.1\\
81	9.1\\
82	9.11\\
83	9.13\\
84	9.18\\
85	9.27\\
86	9.43\\
87	9.6\\
88	9.76\\
89	9.88\\
90	9.94\\
91	9.96\\
92	9.97\\
93	9.99\\
94	9.99\\
95	9.99\\
96	10\\
97	10\\
98	10\\
99	10\\
100	10\\
};
\addplot [color=red,solid,line width=1.5pt,forget plot]
  table[row sep=crcr]{%
0	2\\
1	2.92\\
2	3.38\\
3	3.96\\
4	4.41\\
5	4.68\\
6	4.97\\
7	5.25\\
8	5.51\\
9	5.77\\
10	6.04\\
11	6.26\\
12	6.44\\
13	6.64\\
14	6.74\\
15	6.87\\
16	7.04\\
17	7.11\\
18	7.21\\
19	7.33\\
20	7.42\\
21	7.47\\
22	7.48\\
23	7.53\\
24	7.59\\
25	7.67\\
26	7.73\\
27	7.82\\
28	7.87\\
29	7.95\\
30	8.01\\
31	8.03\\
32	8.09\\
33	8.14\\
34	8.18\\
35	8.24\\
36	8.31\\
37	8.37\\
38	8.42\\
39	8.47\\
40	8.5\\
41	8.52\\
42	8.57\\
43	8.58\\
44	8.62\\
45	8.62\\
46	8.68\\
47	8.73\\
48	8.79\\
49	8.82\\
50	8.84\\
51	8.86\\
52	8.88\\
53	8.89\\
54	8.91\\
55	8.94\\
56	9.02\\
57	9.05\\
58	9.07\\
59	9.09\\
60	9.11\\
61	9.14\\
62	9.17\\
63	9.18\\
64	9.19\\
65	9.22\\
66	9.25\\
67	9.27\\
68	9.29\\
69	9.29\\
70	9.3\\
71	9.31\\
72	9.34\\
73	9.34\\
74	9.35\\
75	9.36\\
76	9.37\\
77	9.37\\
78	9.39\\
79	9.39\\
80	9.41\\
81	9.41\\
82	9.42\\
83	9.42\\
84	9.43\\
85	9.43\\
86	9.44\\
87	9.45\\
88	9.45\\
89	9.45\\
90	9.45\\
91	9.46\\
92	9.47\\
93	9.47\\
94	9.49\\
95	9.49\\
96	9.49\\
97	9.5\\
98	9.5\\
99	9.51\\
100	9.52\\
};

\coordinate (start_arrow_ERMVAR) at (axis cs:17,9.2);
\coordinate (end_arrow_ERMVAR) at (axis cs:15,7.7);

\end{axis}

\draw[->, >=stealth] (start_arrow_ERMVAR) -- (end_arrow_ERMVAR);
\node[anchor=south,inner sep=1pt,font=\scriptsize] (tag) at (start_arrow_ERMVAR) {\textit{GP-Var, ERM}};

\end{tikzpicture}%
    \hfill
    \input{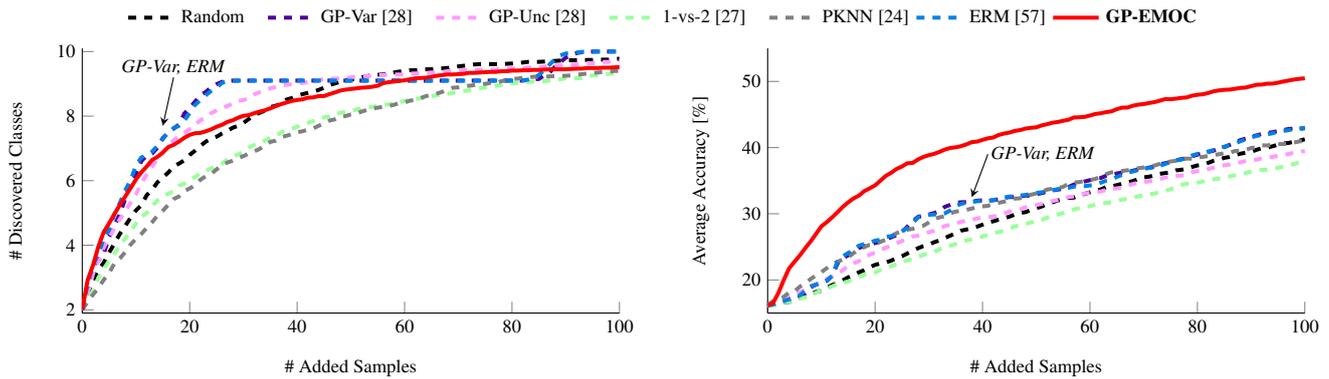}
    \caption{
        Experimental results for active class discovery and improving recognition accuracy with active learning on the COCO dataset~\cite{coco}.
        Baselines are indicated with dotted lines, whereas EMOC is plotted solidly.
    }
    \label{fig:emoc_curves}
\end{figure*}

The following will show the performance of the EMOC criterion on an unsupervised object detection task.
This setup is slightly different to our experiments in \cref{sec:exp,sec:ctrap}.
Here, we do not have access to a detection model which adapts over time. Instead, we rely on object proposals which are generated in an unsupervised manner using a fixed method.
This is different from the previous sections where a complete detection pipeline was trained by optimizing localization and classification jointly.
Hence, the shown pipeline can be seen as an alternative to our current implementation which should provide more insight in possible solutions of the presented problem.

\paragraph{Baselines}
We compare the EMOC approach with the predictive variance (GP-var) as well as uncertainty (GP-unc) of Gaussian processes~\cite{Kapoor10:GP},
the best-vs-second-best strategy (1--vs--2) proposed in \cite{Joshi09:MAL} (also used in \cref{sec:exp}),
the multi-class query strategy based on probabilistic KNN classifiers (PKNN)~\cite{Jain09:ALL}
and the empirical risk minimization approach of \cite{roy2001toward} applied to GP (ERM).
Furthermore, we also include the baseline of random querying.
The EMOC criterion is augmented with the two additives for handling categorical and non-categorical rejections as presented above.
In addition, we also add all rejected examples as negatives to each of the one-vs-all binary classifiers, a strategy that has shown to be valuable also for task adaptation with large-scale datasets~\cite{jia2013latent}.
For a broader evaluation involving more datasets as well as an ablation study see \cite{Kaeding15_ALD}.

\paragraph{Dataset}
For the shown experiment, we use a subset of the COCO training dataset~\cite{coco} and extract object proposals with the geodesic object proposal method of \cite{gop}.
The dataset for our experiment is created as follows:
As a problem domain, we select all animal categories\footnote{\textit{bird}, \textit{cat}, \textit{dog}, \textit{horse}, \textit{sheep}, \textit{cow}, \textit{elephant}, \textit{bear}, \textit{zebra} and \textit{giraffe}.}.
Segments that overlap with more than an intersection-over-union (IoU) value of $0.9$ with a ground-truth object of one of these categories are considered as valid objects and labeled accordingly.
Randomly chosen segments with no overlap with a ground-truth object are considered as unnameable segments, which would be rejected by an annotator.
These segments can be categorical examples (objects of non-animal categories) and non-categorical instances (wrongly detected object proposals).
In total, we use $\numprint{10000}$ random images of the dataset, which contain at least one of the objects of our problem domain.
Thus, we obtain $\numprint{4574}$ boxes showing valid animal instances and $\numprint{3824}$ boxes covering proposals to be rejected.
Features are extracted using outputs of \texttt{pool5}, a layer of a convolutional neural net (CNN) provided by the Caffe framework~\cite{Jia14:CCA} and trained on ImageNet images.
Given the high feature dimensionality, a simple linear kernel is applied.
These features have shown to be powerful for scene understanding tasks, although they have been learned from internet images not related to scenes as contained in the COCO dataset.

\paragraph{Experimental Setup}
In the experiment, we start with an initial set of $2$ known classes and $5$ training examples per class, both randomly selected but identical for each selection method.
We randomly select 10 tasks by splitting classes in known and unknown, and each task is randomly initialized 10 times, resulting in 100 individual test scenarios to average over.
After querying and labeling an example, the classification model is updated and evaluated on a held out test set of 30 examples per class.
Note that in the beginning, the test set also contains examples of classes that are not known to the system since the total number of classes is larger than the number of classes in the initial training set.
All examples that are neither in the test set nor in the initial training set are treated as the unlabeled pool.
This includes all the unnameable proposals.
In all settings, we are interested in fast discovery of all classes as well as high recognition accuracy.

\paragraph{Evaluation}
The experimental results are shown in \cref{fig:emoc_curves}.

In case of the number of discovered classes we can see EMOC to be the fastest in earlier stages of the experiment.
This relates to the \enquote{fast exploration} scenario mentioned in \cref{sec:exp}.
GP-Var, ERM and GP-Unc are able to catch up to EMOC. However, 1-vs-2 and PKNN show a very slow discovery behavior, which is even worse than random selection.
1-vs-2 and GP-Unc also perform worse than random selection in terms of average accuracy.
Interestingly, 1-vs-2 functions well in our previous experiment (see \cref{sec:exp}), where it is the basis for the \emph{Sum}, \emph{Max} and \emph{Avg} methods.
This is possibly due to the type of classifier used or because of the missing interaction with the model responsible for generating proposals.

Finally, all evaluated methods reach roughly the same number of discovered classes after $100$ queries.
The shown curves for average accuracy reveal that even if EMOC could not clearly show an advantage in terms of class discovery in this case (please see \cite{Kaeding15_ALD} for additional experiments),
the selected examples lead to a clear advantage in recognition strength.
Considering both results we can conclude that it is not only necessary to discover as many classes as possible, also how these classes are represented is of high importance.


\section{Conclusion}
In this work, we present a set of methods that efficiently select promising examples to be labeled by a human annotator
for active learning and continuous exploration.
These methods are designed for object detection, with two of the specifically adapted to the popular YOLO method.
We validate the performance of these methods on the PASCAL VOC 2012 benchmark to ensure robustness and accuracy.
The best method is then applied in a real-world scenario where images of camera traps are analyzed for occupancy estimation.
For this application, the viability of active and continuous exploration is demonstrated successfully.
A software implementation of this system, used in the real-world application, is described in detail and available
to researchers upon request.

As an outlook, we also present an active learning method called EMOC that has some theoretical advantages over the heuristics
such as 1-vs-2 that we currently use. As a first step towards including it in our application,
we show that it performs well in a simpler scenario where proposals are generated in an unsupervised manner.
In further work, EMOC could be integrated into a complete detection and localization framework.

\begin{acknowledgements}
We would like to thank Andrea Perino from iDiv for the cooperation, and for supplying wildlife surveillance data as well as annotations.
We also want to thank Alexander Freytag, Erik Rodner and Paul Bodesheim as co-authors of the original publication of multi-class EMOC in \cite{Kaeding15_ALD}.
\end{acknowledgements}

%
%

\bibliographystyle{spmpsci}      
\bibliography{paper}   

\end{document}